\documentclass{nature}

\usepackage{amssymb}
\usepackage{graphicx}
\usepackage{times}
\usepackage{epsfig}
\usepackage{amsmath}
\usepackage{subfigure}
\usepackage{amssymb}
\usepackage{epstopdf}
\graphicspath{ {Pic4/} }

\title{\Large{Object Recognition by Using Multi-level Feature Point Extraction }}

\author{Yang Cheng, Timeo Dubois}

\begin{document}

\maketitle
\section{Abstract}
In this paper, we present a novel approach for object recognition in real-time by employing multi-level feature analysis and demonstrate the practicality of adapting feature extraction into  a Naive Bayesian classification framework that enables simple, efficient, and robust performance. We also show the proposed method scales well as the number of level-classes grows. To effectively understand the patches surrounding a keypoint, the trained classifier uses hundreds of simple binary features and models class posterior probabilities. In addition, the classification process is computationally cheap under the assumed independence between arbitrary sets of features. Even though for some particular scenarios, this assumption can be invalid. We demonstrate that the efficient classifier nevertheless performs remarkably well on image datasets with a large variation in the illumination environment and image capture perspectives. The experiment results show consistent accuracy can be achieved on many challenging dataset while offer interactive speed for large resolution images. The method demonstrates promising results that outperform the state-of-the-art methods on pattern recognition. 

\section{Introduction}
In the literature, one common idea image feature extraction is to focus on finding strong features that is robust enough to deal with perspective changes, lighting variations, such as SIFT
local patterns detection and combines the output of the classifiers. Plagemann et al.6 create a novel
interest point detector for catching body components from depth images.

Identifying textured patches surrounding keypoints across images acquired under widely varying poses and lightning conditions is at the heart of many Computer Vision algorithms \cite{iref1}. The resulting correspondences can be used to register different views of the same scene, extract 3D shape information, or track objects across video frames. Correspondences also play a major role in object category recognition and image retrieval applications \cite{iref2}.

Because of its strength to fractional impediments and computational proficiency, acknowledgment of picture patches removed around distinguished key focuses is critical for some vision issues. Therefore, two fundamental classes of methodologies have been produced to accomplish power to point of view and lighting changes \cite{iref3}. The primary family depends on nearby descriptors intended to be invariant, or possibly hearty, to particular classes of distortions \cite{iref4}.  An inferior depends on measurable learning procedures to figure a probabilistic model of the fix.

\section{A Semi-naive Bayesian approach to patch recognition}

Picture patches can be perceived on the premise of extremely straightforward and arbitrarily picked double tests that are gathered into choice trees and recursively parcel the space of all conceivable patches.
By and by, no single tree is sufficiently discriminative when there are many classes. In any case, utilizing various trees and averaging their votes yields great outcomes in light of the fact that everyone parcels the component space in an unexpected way \cite{iref5}.

\subsection{Formulation of Feature Combination}

Given the patch surrounding a key point detected in an image, our task is to assign it to the most likely class. Let Ci, I= 1, H
Be the set of classes and let fj, j= 1, N
Be the set of binary features that will be calculated over the patch we are trying to classify. Formally, we are looking for

\begin{displaymath}
C_i = arg_Cimax P(C=c_i|f1,f2,...,f_N)
\end{displaymath}

Where C is a random variable that represents the class. Bayes's formula yields

$$P(C=c_i|f1,f2,...,f_N) = \frac{c)P(C=c_i)}{p(f1,f2,...,f_N)}$$

[As the denominator is simply a scaling factor, it can be reduced to]

\begin{displaymath}
C_i = arg_Cimax P(C=c_i|f1,f2,...,f_N)
\end{displaymath}

\[
    f_j=
    \begin{cases}
    1 if I(d_j,_1)  < I(d_j,_2) \\
    0 otherwise
    \end{cases}
\]

Where I represents the image patch \cite{iref7}. (As the features specified are pretty simple, so N~300. So)

$$P(f1,f2,.....,f_N|C=c_i) = \prod_{j=1}^{N} P(f_j|C=c_i)$$

\section{Comparison With Randomized Trees}

\begin{figure*}[!htb]
\centerline{\epsfig{figure=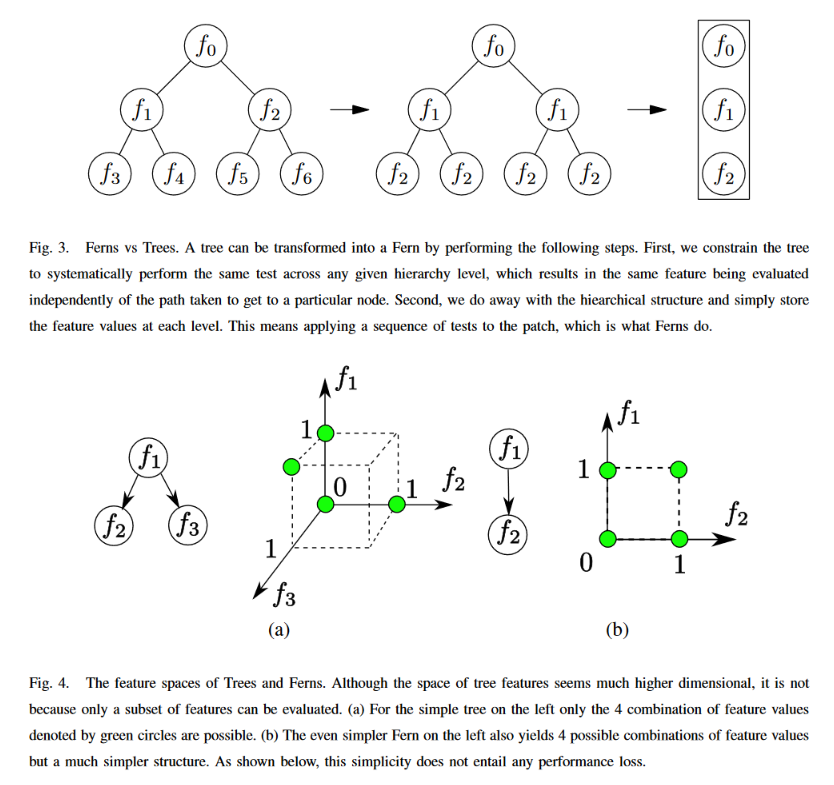, width=11cm}}
\end{figure*}
As shown in Figures 3 and 4, Ferns can be considered as simplified trees. To compare RTs and Ferns, we experimented with the three images of Figure 5.

\begin{figure*}[!htb]
\centerline{\epsfig{figure=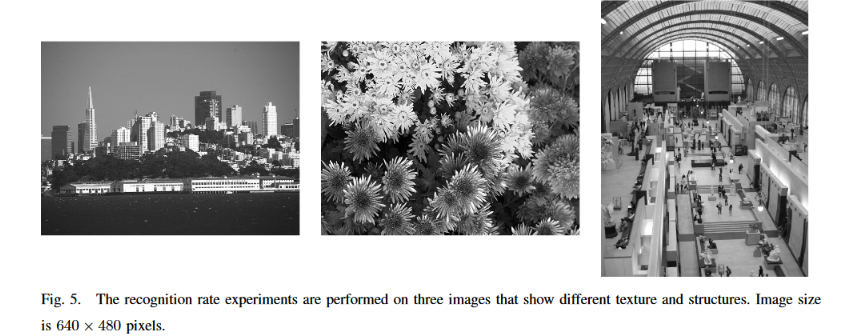, width=11cm}}
\end{figure*}
Greeneries vary from trees in two vital regards: The probabilities are increased in a Naive-Bayesian manner as opposed to being arrived at the midpoint of and the various leveled structure is supplanted by a level one.

The preparation set is acquired by arbitrarily twisting pictures of Figure 5. To play out these examinations, we speak to relative picture disfigurements as 2×2 networks of the shape.
$$R_0R_-\varnothing diag(\lambda1,\lambda2)R_\varnothing$$

Where diag (?1, ?2) is a corner to corner 2 × 2 framework and R? speaks to a revolution of point ? . Both to prepare and to test our greeneries, we distorted the first pictures utilizing such misshapenness registered by arbitrarily picking ? and f in the [0: 2p] territory and ?1 and ?2 in the [0.6: 1.5] territory. Fig. 6 delineates patches encompassing individual intrigue focuses first in the first pictures and after that in the distorted ones \cite{iref7}. We utilized 30 irregular relative distortions for each level of pivot to create 10800 pictures.

For the most part the test set is acquired by creating a different arrangement of 1000 pictures in a similar relative twisting extent and including clamor. In Figure 7, we plot the outcomes as an element of the quantity of trees or Ferns being utilized.

\begin{figure*}[!htb]
\centerline{\epsfig{figure=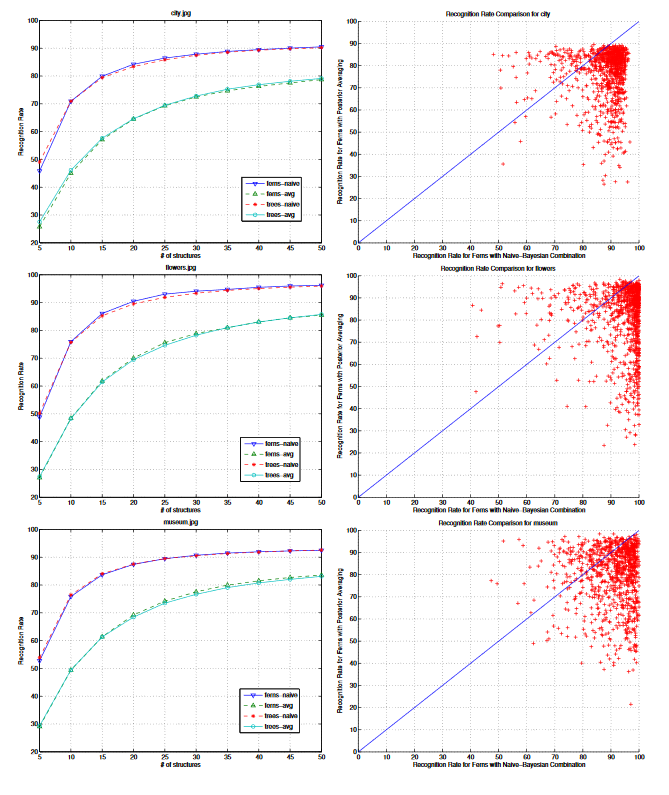, width=11cm}}
\end{figure*}
We first note that utilizing either level Fern or progressive tree structures does not influence the acknowledgment rate, which was not out of the ordinary as the components are taken totally at irregular. Moreover as the scramble plots of Figure 7 appear, for the Naive-Bayesian blend the acknowledgment rate on individual twisted pictures never falls beneath a worthy rate.

\section{Experiments}

It is hard to play out a totally reasonable speed examination between our Ferns and SIFT for a few reasons \cite{iref16}. Filter reuses moderate information from the key guide extraction toward register canonic scale and introductions and the descriptors, while greeneries can depend on a minimal effort key-point extraction \cite{iref8}.
                                Ferns vs SIFT to detect 3D objects
So far we have considered that the key-focuses lie on a planar protest and assessed the heartiness of Ferns concerning viewpoint impacts. This rearranges preparing as a solitary view is adequate and the known 2D geometry can be utilized to figure ground truth correspondences. However most protests have really three dimensional appearance, which suggests that self-impediments and complex enlightenments impacts must be considered to effectively assess the execution of any key-point coordinating calculation \cite{iref9}.

\begin{figure*}[!htb]
\centerline{\epsfig{figure=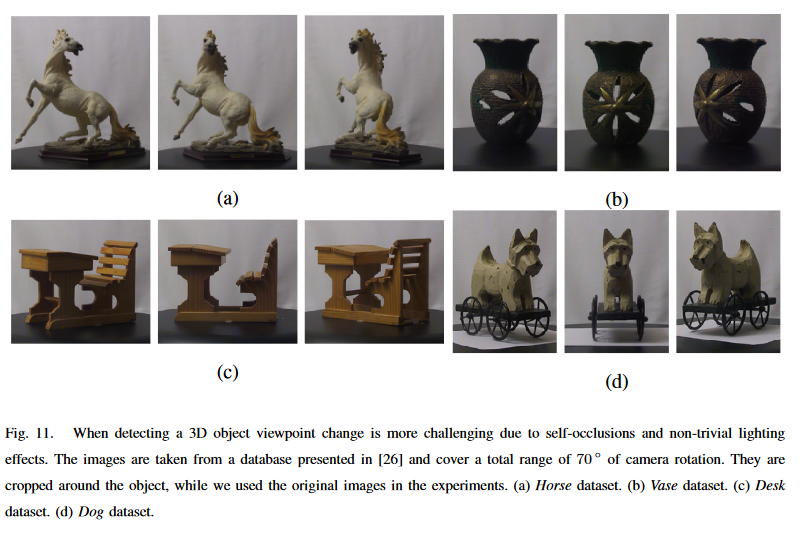, width=11cm}}
\end{figure*}
Figure 11 shows such pictures traversing a 70? camera revolution go. We utilized this picture database to assess the execution of Ferns for an assortment of 3D items. We look at our outcomes against the SIFT identifier/descriptor match which has been found to perform extremely well on this database. The key-focuses and the descriptors are registered utilizing an indistinguishable programming from some time recently. We acquired the ground truth by utilizing simply geometric strategies, which is conceivable on the grounds that the cameras and the turn table are adjusted \cite{iref10}. The underlying correspondences are acquired by utilizing the trifocal geometry between the top/base cameras in the inside view and each other camera as outlined by Figure 12 \cite{iref15}. We then reproduce the 3D focuses for each such correspondence in the base/focus camera arrange casing and utilize these to frame the underlying tracks that traverse the-35?/+35?rotation territory around a focal view.
\begin{figure*}[!htb]
\centerline{\epsfig{figure=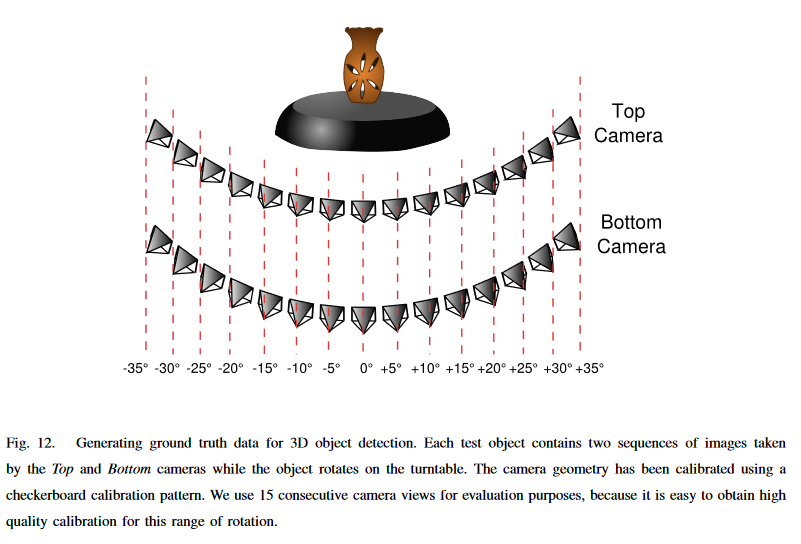, width=11cm}}
\end{figure*}
At last to build power against spurious tracks framed by anomalies, we take out tracks covering less than 30percent of the perspectives and the rest of the tracks shape the ground truth for the assessment \cite{iref13}, which is free of exceptions. Test ground truth information is portrayed by Figure 13, which demonstrates the mind boggling varieties in fix appearance incited by the 3D structure of the items \cite{iref11}.
\begin{figure*}[!htb]
\centerline{\epsfig{figure=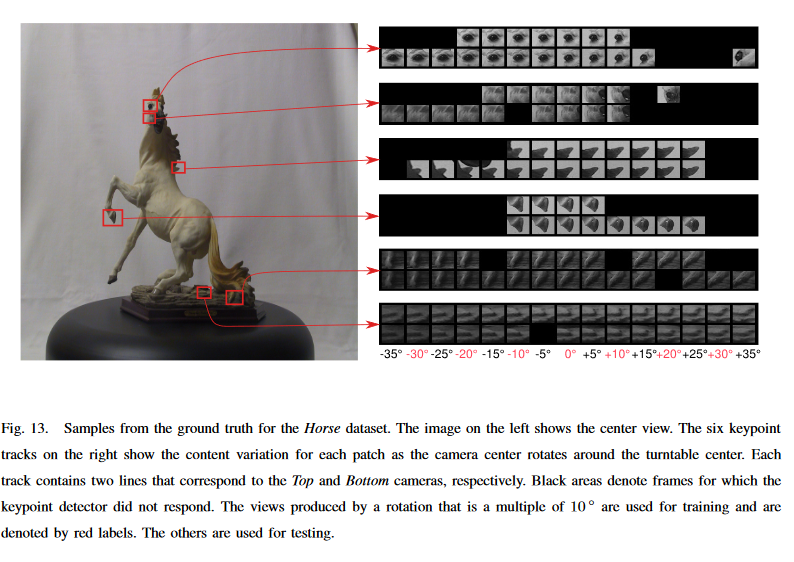, width=11cm}}
\end{figure*}

\section{Panorama and 3–D Scene Annotation}
With the current expansion of cell phones with huge handling power, there has been a surge of enthusiasm for building genuine applications that can consequently explain the photographs and give helpful data about spots of intrigue.
\begin{figure*}[!htb]
\centerline{\epsfig{figure=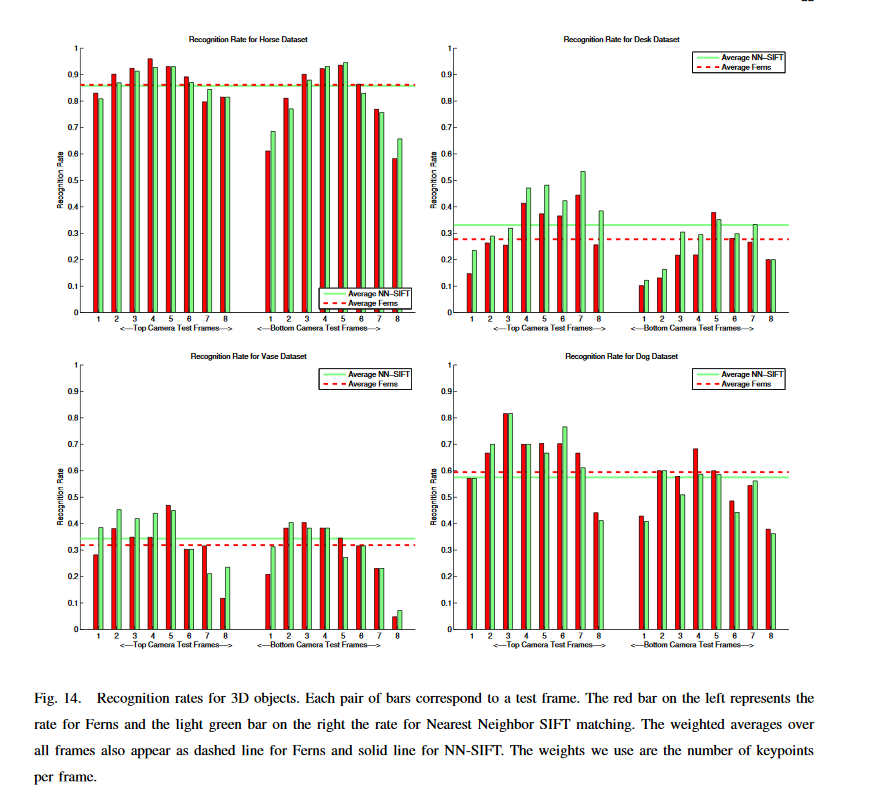, width=11cm}}
\end{figure*}
We have tried Ferns on two such applications, comment of display scenes and parts of a verifiable working with 3–D structure \cite{iref14}. Both applications run easily at casing rate utilizing a standard portable PC and an of the rack web camera. By applying standard improvements for implanted equipment, we have ported this execution onto a cell phone that keeps running at a couple outlines for every second.

\section{Conclusion}
We have introduced an intense strategy for picture fix acknowledgment that performs well even within the sight of extreme viewpoint twisting.  A key part of our approach is the Naive-Bayesian blend of classifiers that obviously beats the averaging of probabilities we utilized as a part of prior work. We have demonstrated that such a credulous mix methodology is a beneficial option when the particular issue is not excessively delicate to the suggested autonomy suspicions.

%% Here is the endmatter stuff: Supplementary Info, etc.
%% Use \item's to separate, default label is "Acknowledgements"

%%
%% TABLES
%%
%% If there are any tables, put them here.
%%

\end{document}